\documentclass{article}

\usepackage{PRIMEarxiv}

\usepackage[utf8]{inputenc} 
\usepackage[T1]{fontenc}    
\usepackage{hyperref}       
\usepackage{url}            
\usepackage{booktabs}       
\usepackage{amsfonts}       
\usepackage{nicefrac}       
\usepackage{microtype}      
\usepackage{lipsum}
\usepackage{fancyhdr}       
\usepackage{amsmath,amssymb,amsfonts}
\usepackage{algorithmic}
\usepackage{booktabs}
\usepackage{tabularx}
\usepackage{multirow}
\usepackage{textcomp}
\usepackage{tabularray}
\usepackage{graphicx}       
\graphicspath{{media/}}     

\title{Crossed-IoT device portability of Electromagnetic Side Channel Analysis: Challenges and Dataset

}

\author{
  Tharindu Lakshan Yasarathna \\
  School of Computer Science \\
  University College Dublin \\
  Belfield, Dublin, Dublin 4, Ireland\\
  \texttt{} \\
   \And
 Lojenaa Navanesan \\
  School of Computing \\
  University of Colombo \\
  Sri Lanka\\
  \texttt{} \\
   \And
  Simon Barque \\
    School of Computer Science \\
  University College Dublin \\
  Belfield, Dublin, Dublin 4, Ireland\\
  \texttt{} \\
  \And
 Assanka Sayakkara \\
  School of Computing \\
  University of Colombo \\
  Sri Lanka\\
  \texttt{} \\
  \And
  Nhien-An Le-Khac \\
    School of Computer Science \\
  University College Dublin \\
  Belfield, Dublin, Dublin 4, Ireland\\
  \texttt{} \\
}

\begin{document}
\maketitle

\begin{abstract}
IoT (Internet of Things) refers to the network of interconnected physical devices, vehicles, home appliances, and other items embedded with sensors, software, and connectivity, enabling them to collect and exchange data. IoT Forensics is collecting and analyzing digital evidence from IoT devices to investigate cybercrimes, security breaches, and other malicious activities that may have taken place on these connected devices. In particular, Electromagnetic Side-Channel Analysis (EM-SCA) has become an essential tool for IoT forensics due to its ability to reveal confidential information about the internal workings of IoT devices without interfering these devices or wiretapping their networks. However, the accuracy and reliability of EM-SCA results can be limited by device variability, environmental factors, and data collection and processing methods. Besides, there is very few research on these limitations that affects significantly the accuracy of EM-SCA approaches for the crossed-IoT device portability as well as limited research on the possible solutions to address such challenge. Therefore, this empirical study examines the impact of device variability on the accuracy and reliability of EM-SCA approaches, in particular machine-learning (ML) based approaches for EM-SCA. We firstly presents the background, basic concepts and techniques used to evaluate the limitations of current EM-SCA approaches and datasets. Our study then addresses one of the most important limitation, which is caused by the multi-core architecture of the processors (aka. System-On-Chip). We present an approach to collect the EM-SCA datasets and demonstrate the feasibility of using transfer learning to obtain more meaningful and reliable results from EM-SCA in IoT forensics of crossed-IoT devices. Our study moreover contributes a new dataset for using deep learning models in analysing Electromagnetic Side-Channel data with regards to the cross-device portability matter.
\end{abstract}


\section{Introduction}
\label{sec:introduction}
The Internet of Things (IoT) is a rapidly expanding technology field that is increasingly becoming a part of our daily lives. IoT devices are connected to the internet and can interact with other devices, people, and systems~\cite{gokhale2018introduction}. With the continuous growth of the number of IoT devices, it needs a more secure and efficient way of managing the data being collected, transmitted, and processed. IoT forensics~\cite{atlam2020internet} has emerged as a critical aspect of this field used to investigate and prevent malicious activities on IoT devices~\cite{conti2018internet}. In particular, Electromagnetic Side-Channel Analysis (EM-SCA) has become a promising approach for IoT forensics due to its ability to reveal confidential information about the internal workings of IoT devices without interfering these devices. 
Electromagnetic Side-Channel Analysis (EM-SCA) is a method of analyzing the electromagnetic emanations that are produced by electronic devices. These emanations can provide information about the device's internal workings, including power consumption, encryption key, and processing speed~\cite{ghosh2022electromagnetic}. This information can be used to perform a range of forensic activities, including data recovery, malware analysis, and key extraction. In the context of IoT forensics, EMSCA can extract sensitive information from IoT devices, such as login credentials, private keys, and other confidential information~\cite{sayakkara2020facilitating}. By analyzing the electromagnetic emanations produced by IoT devices, it is possible to gain insight into the device's internal operations and identify potential security vulnerabilities~\cite{sayakkara2019survey}.

On the other hand, artificial intelligent (AI) analytical techniques with machine learning (ML) and deep learning (DL) approaches are becoming more sophisticated and powerful. ML algorithms have been applied successfully for a variety of application domains such as agriculture~\cite{KAMILARIS201870}, renewable energies~\cite{SriRevathi2023}, climate~\cite{ZENNARO2021103752}, and health systems ~\cite{Bhardwaj2017}. Moreover, ML/DL techniques have been widely used in cybersecurity~\cite{Shaukat2020} and digital forensics domains~\cite{Shahzad2020} to assist the investigators to fight against cybercrimes. Recently, ML/DL techniques have also used in EM-SCA to enhance the capability of identifying application's activities of mobile devices~\cite{Sayakkara2021smartphone} and of IoT devices ~\cite{sayakkara2021electromagnetic} to detect required artefacts as well as potential security issues. Relevant EM-SCA datasets have been created for these purposes~\cite{Sayakkara2021smartphone}~\cite{sayakkara2021electromagnetic}. 

Today ML/DL approaches for EM-SCA have become a new essential research trend in cybersecurity and digital forensic domains, especially in the cross-device portability problems where ML/DL models of EM-SCA of activities trained from a given device could be used to detect relevant activities in other devices of the same kind.

However, despite its usefulness, the accuracy and reliability of ML/DL models for EM-SCA depends on the EM datasets' quality that can make these models more difficult to obtain meaningful insights from the data. Some of the limitations of EMSCA datasets can be listed as follows: 
 
\begin{itemize}
  \item Device variability: The electromagnetic emanations produced by different devices can vary widely, affecting the results' accuracy. 
  
  \item Environmental factors: The electromagnetic environment can also impact the results, affecting the strength and quality of the electromagnetic emanations. 
  
  \item Data collection and processing: The quality and accuracy of the data collected and processed by EM-SCA are also critical factors that can affect the reliability of the results. 
  
\end{itemize}

Among these limitations, when having the electromagnetic (EM) data captured and processed in the same manner, the device variability becomes the most important factor that affects the overall accuracy of ML/DL models for EM-SCA. This is more severe when EM data of the same activities captured from the different devices of the same kind are also different where are they should be at certain levels of similarity~\cite{sayakkara2021electromagnetic}, given these devices are running the same SoC (system-on-chip). This issue also limits the ability of applying ML/DL-based approaches for EM-SCA in the real-world applications. To the best of our knowledge, there is no research in literature that studied this limitation properly in terms of how and why EM data captured could be dissimilar from devices of the same kind and how to overcome this issue for ML/DL approaches.

Therefore, in this paper, we firstly study challenges of using ML/DL based model for EM-SCA in the context of crossed-IoT device portability. We then examine the ability of using transfer learning to address these issues. Finally, we also create and validate a new EM-SCA dataset, which is ready for building transfer learning models for EM-SCA with regards to the crossed-IoT device portability issues. This dataset can be exploited in both cybersecurity and digital forensic domains. This paper makes the following contributions to the domain of IoT Forensics:

\begin{itemize}
    \item Examination of the impact of device variability, environmental factors, and data collection and processing on the accuracy and reliability of EM-SCA results.

    \item Analysis the feasibility of using transfer learning in improving the performance of ML/DL models used in analyzing EM-SCA data.

    \item Discussion of the importance of addressing these limitations in order to obtain more meaningful and reliable results from EM-SCA in IoT forensics.

    \item A new public dataset for using deep learning models in analysing EM-SCA data with regards to the cross-device portability matter\footnote{\url{https://aseados.ucd.ie/datasets/EMSCA-2023-Latest/}}.

\end{itemize}

In the following section, we discuss some of the existing state-of-art research related to our topic. In Section \ref{sec:methodology}, we present our methodology with a detailed explanation followed by experimental analysis in Section \ref{sec:experimentalanalysis}. Section \ref{sec:discussion} is about the discussion of our findings. Finally in Section \ref{sec:conclusion}, we conclude this paper by addressing some open questions to the research community and, our future direction of the research. 

\section{Related Work}
\label{sec:relatedwork}
In recent years, there has been a growing interest in IoT forensics, particularly in EM-SCA. Researchers and practitioners have been exploring the potential of EM-SCA to analyze the electromagnetic emanations produced by IoT devices to reveal information about the device's internal operations. In this section, we will review some recent studies conducted in the field of EM-SCA for IoT forensics.

One of the earliest studies in this area was conducted by~\cite{hong2017extracting}, who used EM-SCA to extract the encryption key from an IoT device. They found that the electromagnetic emanations produced by the device contained information about the encryption key, which could be used to decrypt the data transmitted by the device. This study demonstrated the potential of EM-SCA as a tool for IoT forensics and highlighted the need for better security measures in IoT devices to protect against attacks. Another study by~\cite{kim2018detecting} focused on using EM-SCA to detect malware in IoT devices. The authors used EM-SCA to analyze the electromagnetic emanations produced by a range of IoT devices and found that the emanations could reveal information about malware in the machine. They also found that the results were consistent with other malware detection methods, demonstrating the potential of EM-SCA as a complementary tool for detecting malware in IoT devices.

A study by~\cite{zhao2019limitations} investigated the limitations of EM-SCA in IoT forensics by examining the variability of the electromagnetic emanations produced by different devices. They found that the electromagnetic emanations produced by different devices could vary widely, affecting the accuracy of the results obtained from EM-SCA. The authors also found that the variability was influenced by various factors, including the device's operating system, hardware, and environmental conditions. This study highlights the importance of considering the variability of electromagnetic emanations when using EM-SCA for IoT forensics.
More recently, a study by~\cite{chen2020machine},~\cite{levina2016side} explored machine learning techniques in EM-SCA for IoT forensics. The authors found that ML algorithms could be used to improve the accuracy and reliability of the results obtained from EM-SCA. They also found that using ML algorithms could reduce the dependence on large and representative datasets, making EM-SCA more accessible and practical for IoT forensics.

EM-SCA is a powerful tool for IoT forensics that has the potential to reveal confidential information about the internal workings of IoT devices. Recent studies have demonstrated the potential of EM-SCA as a tool for extracting encryption keys, detecting malware, and improving the accuracy and reliability of the results obtained from EM-SCA. However, some limitations and challenges need to be addressed to fully realize the potential of EM-SCA for IoT forensics, such as the variability of the electromagnetic emanations produced by different devices, the dependence on large and representative datasets, and the need for improved data collection and processing methods. Therefore, further research is needed to address these limitations and challenges and improve IoT forensics' overall quality using EM-SCA.

\section{Adopted Approach}
\label{sec:methodology}

\begin{figure}[!ht]
    \centering
    \includegraphics[width= 8.5cm]{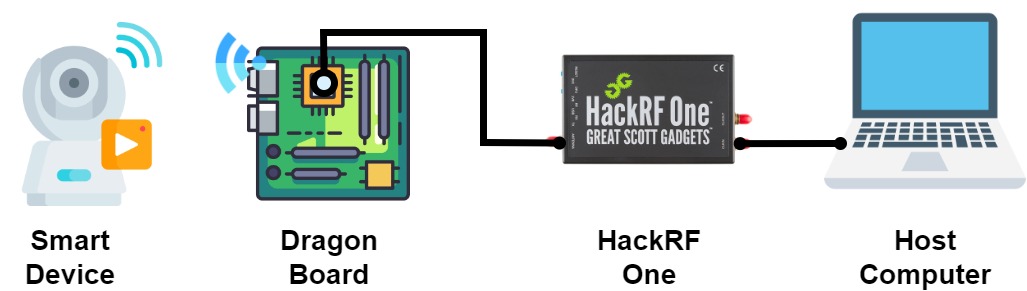}
    \caption{The hardware setup for acquiring EM radiation from Dragonboard. In the meantime, it connects with the smart device through the custom-built WiFi network.}
    \label{fig:hwsetup}
\end{figure}

Although ML/DL approaches for EM-SCA have become a new essential research trend in cybersecurity and digital forensic domains, the current approaches in literature are mostly working on the datasets generated from the same device. The problem of crossed-IoT device portability has not been studied in details. It limits the ability of deploying ML/DL approaches for EM-SCA in the real-world applications of cybersecurtiy and digital forensics. One of the challenges is the dropping significantly of ML/DL models' accuracy when applying in the crossed-IoT device context. To address this challenge, in this paper we focus on three main research questions about the possible factors that affects the ML/DL models' accuracy and how to improve it. These research questions are listed as follows:

\begin{itemize}
    \item \textbf{R1}: How the multi-core architecture of SoC integrated on these devices affects the ML/DL models' accuracy?
    \item \textbf{R2}: How the number of activities running on the IoT device affect the ML/DL models' accuracy?
    \item \textbf{R3}: Can transfer learning techniques be applied to improve the ML/DL models' accuracy?
\end{itemize}

We address these research questions through our empirical studies where we set up relevant experiments that we describe in this section. We then deploy intensively tests in varied scenarios and the analysis of the results will be described in the following section.

This section describes the necessary technological background for those features, along with the experimental platform and data generation steps.

\subsection{Experimental Platform}
\begin{figure}[ht]
    \centering
    \includegraphics[width= 12cm]{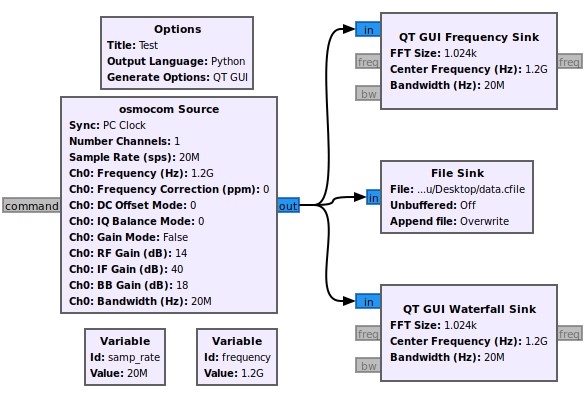}
    \caption{The GNU Radio Companion (GRC) flow graph for acquiring EM traces.}
    \label{fig:osmo}
\end{figure}

Acquiring EM radiation from a computing system requires multiple hardware and software components. In this work, we used Dragonboard connected to a smart device through the It's WiFi as a device-under-test (DUT)~\cite{sayakkara2021electromagnetic}. Apart from that, signal-capturing equipment (HackRF One) is used with an antenna to capture the EM radiation of the DUT (cf. Figure~\ref{fig:hwsetup}). The signal acquisition equipment is connected to a host computer that runs the necessary software to read the EM data samples and save them into trace files (cf. Figure~\ref{fig:osmo}).

\begin{table*}[ht]
\centering
\caption{Technical specifications of the smart devices used to create the EM dataset.}
\label{tab:devices}
\resizebox{\textwidth}{!}{%
\def\arraystretch{1.3}
\begin{tabular}{cccll}
\hline
\textbf{\begin{tabular}[c]{@{}c@{}}Device\\  x \\ Number \\ of devices\end{tabular}} &
  \textbf{Processor} &
  \textbf{\begin{tabular}[c]{@{}c@{}}CPU \\ Frequency\end{tabular}} &
  \multicolumn{1}{c}{\textbf{WiFi Bands}} &
  \multicolumn{1}{c}{\textbf{Scenarios used to collect data}} \\ \hline
\begin{tabular}[c]{@{}c@{}}DragonBoard \\ 410c \\ x 2 devices\end{tabular} &
  \begin{tabular}[c]{@{}c@{}}Qualcomm® \\ APQ8016e\end{tabular} &
  \begin{tabular}[c]{@{}c@{}}1.2 GHz \\ (4 core)\end{tabular} &
  \begin{tabular}[c]{@{}l@{}}On-board Wi-Fi, \\ Wi-Fi 802.11/g \\ 2.4GHz\end{tabular} &
  \begin{tabular}[c]{@{}l@{}}(\textbf{EXP1}) Connect one smart device to built WiFi \\ network and collect EM traces from the CPU. \\ (\textbf{EXP2}) Consider Dragon Board as DUT and run \\separate Print, Math, Memory, and I/O activities \\on it. Here we use only one core (use the same \\core for all activities) and collect CPU EM traces.\end{tabular} \\ 
\begin{tabular}[c]{@{}c@{}}Amazon Echo \\ Show 5\\ x 2 devices\end{tabular} &
  \begin{tabular}[c]{@{}c@{}}MediaTek \\ MT 8163\end{tabular} &
  \begin{tabular}[c]{@{}c@{}}1.5 GHz \\ (4 core)\end{tabular} &
  \begin{tabular}[c]{@{}l@{}}Dual-band Wi-Fi \\ 802.11 a/b/g/n/ac \\ 2.4 GHz and 5 GHz\end{tabular} &
  \begin{tabular}[c]{@{}l@{}}(1) asking for time, (2) asking to play music, (3) \\ asking for the weather today,(4) asking to play \\ NRJ radio\end{tabular} \\ \hline
\end{tabular}%
}
\end{table*}

For this purpose, DragonBoard 410c Development Board is used based on a Qualcomm APQ8016e application processor with a 1.2 GHz frequency (cf. Table \ref{tab:devices}). Linaro Linux distribution is used as the operating system of DragonBoard and configured as a Routed Wireless Access Point. The new wireless network with the SSID~\textit{DB} used to connect smart devices to Dragonboard. Although, a HackRF One SDR is used as the EM radiation acquisition equipment. The device has a maximum sampling rate of 20 MHz. It supports a data acquisition frequency range from 1 MHz to 6 GHz, and the H-loop antenna is connected to HackRF One to acquire data from the DUT within close proximity. GNU Radio library is employed on the host computer to configure the SDR device and process the data it produces~\cite{sayakkara2021electromagnetic}. 

\begin{figure}[ht]
    \centering
    \includegraphics[width= 12cm]{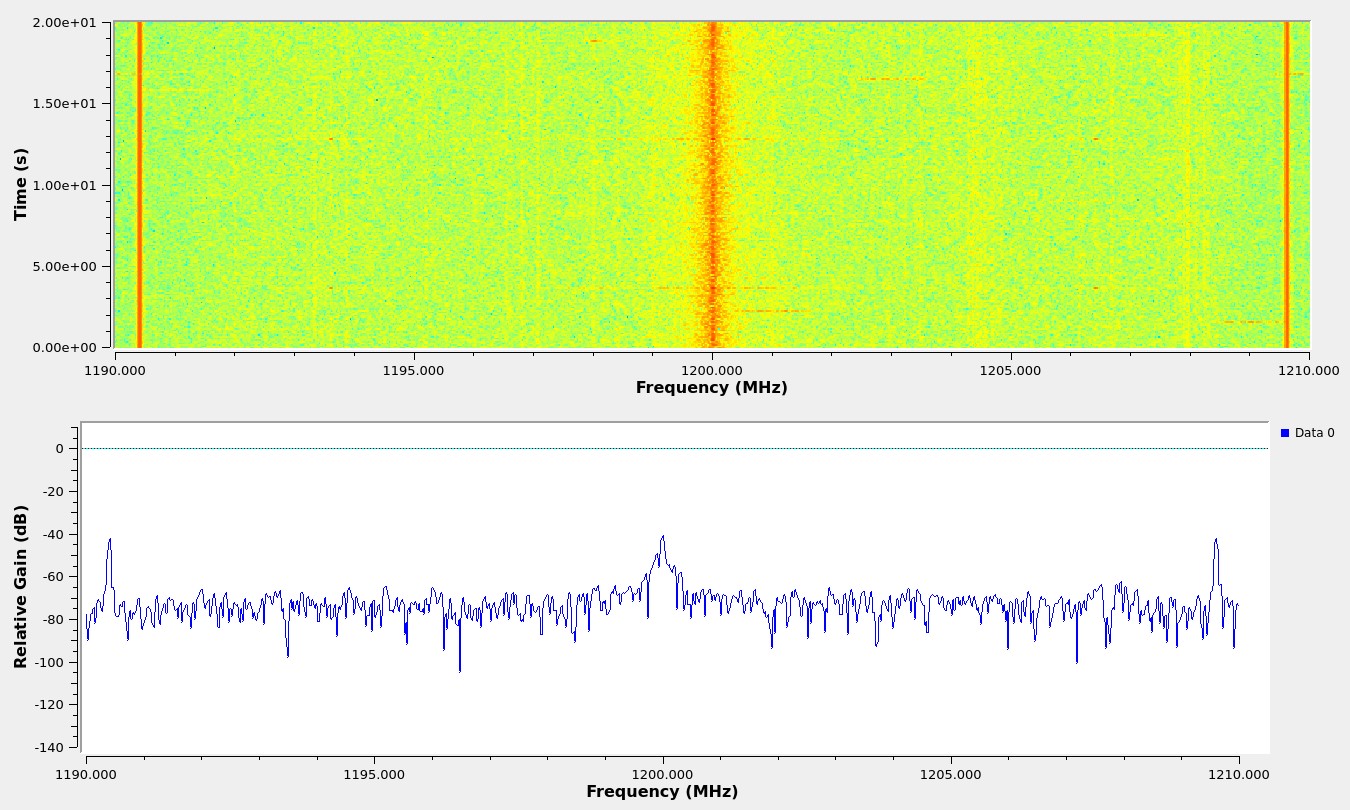}
    \caption{The GNU Radio library visualisation of EM radiation generating from
Dragonboard at 1.2 GHz processor clock frequency, captured with a
sample rate of 20 MHz using an HackRF One.}
    \label{fig:osmoviz}
\end{figure}

The GNU Radio library provides a graphical interface called GNU Radio Companion (GRC), which facilitates creating visual flow graphs to build EM data processing pipelines. To capture EM traces from DUT, place the H-loop antenna on the optimal position of the CPU of the DragonBoard using the GNU Radio library visualizations. The flow diagram created by GNU Radio Companion to record EM traces is shown in Figure~\ref{fig:osmo}. Osmocom Source depicts how the HackRF SDR device is set up to provide I/Q data samples. While the File Sink records the I/Q data stream into a raw data file, the Frequency Sink and Waterfall Sink are used to visualize data (see Figure~\ref{fig:osmoviz}). 

\subsection{Electromagnetic Dataset Generation}
To investigate the device variability, we used Amazon Echo Show 5 and Dragon Board 410c for the experiments (cf. Table \ref{tab:devices}). We used two (02) similar devices to create a training dataset from one device and another to create a testing dataset. The sample rate of the HackRF One SDR was set to 20 MHz, which is the device's maximum capacity. In experiment 1 (\textbf{EXP1}), capture EM traces from the Dragon Board CPU while the smart device is connected through our custom WiFi network~\textit{DB}. In experiment 2 (\textbf{EXP2}), we run different activities on Dragon Board (without connecting other smart devices) utilizing single core (core 1) and capture EM traces from it. We do the same for all devices individually. For this purpose, we need to set the center frequency to HackRF One SDR and put it as 1.2GHz according to the Dragon Board processor clock frequency. Using the pre-described experimental setup, we capture EM traces from each device for the activities shown in Table~\ref{tab:devices}. We also disabled Frequency Sink and Waterfall Sink when File Sink was writing data into a raw data file with \textit{.cfile} extension in the host computer. Two sampling methods are available for analyzing EM radiation captured through signal acquisition equipment, 1) real-valued sampling and 2) complex In-phase and Quadrature-phase (I/Q) sampling~\cite{sayakkara2021electromagnetic}. Due to the high expense of data-collecting equipment and the overhead of storing and processing massive amounts of data, we used complex In-phase and Quadrature-phase (I/Q) sampling method. Build separate data files for each device activity shown in Table~\ref{tab:devices} and combine them to build the final dataset for each device in the latter stage (cf. Table~\ref{tab:dataset}). Also, we keep each data file size under 3GB due to processing limitations.

\begin{table}[ht]
\centering
\caption{Statistics of Datasets}
\label{tab:dataset}
{%
\def\arraystretch{1.2}
\begin{tabular}{lccc}
\hline
\multicolumn{1}{c}{\textbf{Device}} &
  \textbf{\begin{tabular}[c]{@{}c@{}}Number \\ of classes\end{tabular}} &
  \textbf{\begin{tabular}[c]{@{}c@{}}Samples size \\ per class\end{tabular}} &
  \textbf{Total data samples} \\ \hline
Dragon Board 1    & 4 & 30,000 & 120,000 \\ 
Dragon Board 2    & 4 & 30,000 & 120,000 \\ 
Echo Show 1   & 4 & 30,000 & 120,000 \\ 
Echo Show 2   & 4 & 30,000 & 120,000 \\ \hline

\end{tabular}%
}
\end{table}

\section{Experiments and Findings}
\label{sec:experimentalanalysis}
ML methods are particularly well-suited for analyzing large and complex datasets and make predictions. But the EM trace data can not be directly used for ML methods and must be pre-processed to extract features. Short-time Fourier transform (STFT) is used to accomplish this task~\cite{sayakkara2021electromagnetic}. In the experiments, we used 4096 I/Q samples as Fast Fourier Transform(FFT) size and an FFT overlap of 512 I/Q samples. The STFT converted dataset belongs to the label of the original smart device’s software activity and it can be used for machine learning tasks.

From each class, 30,000 samples were used to build the final datasets for all devices. In contrast, MinMaxScaler is used to normalize the data that can be used to transform numerical data to a standard scale, which can improve algorithm performance, data visualization, and comparability of different datasets. In the current analysis, one data set from each device was first split into training and testing with the ratio of 70:30. Keras sequential model was used to build a deep neural network with six (06) hidden layers and trained using 30 epochs with 10\% of the validation set. Keep the created model and test it with the complete data set from another similar device.

The accuracy, precision, recall, and F1 scores~\cite{sayakkara2021electromagnetic} were used to evaluate the performance of the models. We used a previous study~\cite{sayakkara2021electromagnetic} as a baseline to assess the results obtained from our tests on Dragon Board and Echo Show devices. This study was quite similar to ours in that it also involved capturing electromagnetic traces. In this study, authors achieved 99.66\% accuracy with better precision, recall, and F1 score for all classes of the Echo Show device. Although, the authors of study~\cite{sayakkara2021electromagnetic} captured EM traces from other IoT devices like Google Home, Samsung Smarthings Hub, and several Smart Phones from different vendors. Then they analyzed and built the machine-learning models. Deep neural network-based models built using those EM traces captured from each smart device archived an accuracy of over 98.0\%. In our experiments on Dragon Board, we use this observation as the baseline for model accuracy of 98.0\%. 

\begin{table*}[ht]
\centering
\caption{Classification report of the sequential Keras model for Amazon Echo Show device 1. Data set was split with a 70:30 ratio to training (70\%) and testing (30\%) the model ( 0-asking to play music, 1-asking to play radio, 2-asking for time, 3-asking for weather today ).}
\label{tab:echoshow}
\resizebox{\textwidth}{!}{%
\def\arraystretch{1.2}
\begin{tabular}{ccccccccc}
\hline
 &
  \multicolumn{4}{c}{\textbf{Training Data}} &
  \multicolumn{4}{c}{\textbf{Testing Data}} \\ \hline
\textbf{label} &
  \multicolumn{1}{c}{\textbf{precision}} &
  \multicolumn{1}{c}{\textbf{recall}} &
  \multicolumn{1}{c}{\textbf{F1-score}} &
  \textbf{accuracy} &
  \multicolumn{1}{c}{\textbf{precision}} &
  \multicolumn{1}{c}{\textbf{recall}} &
  \multicolumn{1}{c}{\textbf{F1-score}} &
  \textbf{accuracy} \\ \hline
0 &
  \multicolumn{1}{c}{0.96} &
  \multicolumn{1}{c}{0.83} &
  \multicolumn{1}{c}{0.89} &
  \multirow{4}{*}{0.91} &
  \multicolumn{1}{c}{0.95} &
  \multicolumn{1}{c}{0.80} &
  \multicolumn{1}{c}{0.87} &
  \multirow{4}{*}{0.89} \\ 
1 &
  \multicolumn{1}{c}{1.00} &
  \multicolumn{1}{c}{1.00} &
  \multicolumn{1}{c}{1.00} &
   &
  \multicolumn{1}{c}{1.00} &
  \multicolumn{1}{c}{1.00} &
  \multicolumn{1}{c}{1.00} &
   \\ 
2 &
  \multicolumn{1}{c}{0.95} &
  \multicolumn{1}{c}{0.88} &
  \multicolumn{1}{c}{0.92} &
   &
  \multicolumn{1}{c}{0.94} &
  \multicolumn{1}{c}{0.86} &
  \multicolumn{1}{c}{0.90} &
   \\ 
3 &
  \multicolumn{1}{c}{0.77} &
  \multicolumn{1}{c}{0.93} &
  \multicolumn{1}{c}{0.84} &
   &
  \multicolumn{1}{c}{0.73} &
  \multicolumn{1}{c}{0.90} &
  \multicolumn{1}{c}{0.81} &
   \\ \hline
\end{tabular}%
}
\end{table*}

\begin{table}[ht]
\centering
\caption{Classification report of the sequential Keras model for Dragon Board 1 for four (04) activities. The data set was split with a 70:30 ratio to training (70\%) and testing (30\%) of the model ( 0-Print Task, 1-Math Task, 2-Memory Task, 3- I/O Task)}
\label{tab:DB1_4 activity}
\resizebox{\textwidth}{!}{%
\def\arraystretch{1.2}
\begin{tabular}{ccccccccc}
\hline
 &
  \multicolumn{4}{c}{\textbf{Traning Data}} &
  \multicolumn{4}{c}{\textbf{Testing Data}} \\ \hline
\textbf{label} &
  \multicolumn{1}{c}{\textbf{precision}} &
  \multicolumn{1}{c}{\textbf{recall}} &
  \multicolumn{1}{c}{\textbf{F1-score}} &
  \textbf{accuracy} &
  \multicolumn{1}{c}{\textbf{precision}} &
  \multicolumn{1}{c}{\textbf{recall}} &
  \multicolumn{1}{c}{\textbf{F1-score}} &
  \textbf{accuracy} \\ \hline
0 &
  \multicolumn{1}{c}{0.97} &
  \multicolumn{1}{c}{0.98} &
  \multicolumn{1}{c}{0.97} &
  \multirow{4}{*}{0.96} &
  \multicolumn{1}{c}{0.96} &
  \multicolumn{1}{c}{0.98} &
  \multicolumn{1}{c}{0.97} &
  \multirow{4}{*}{0.95} \\ 
1 &
  \multicolumn{1}{c}{0.98} &
  \multicolumn{1}{c}{0.96} &
  \multicolumn{1}{c}{0.97} &
   &
  \multicolumn{1}{c}{0.97} &
  \multicolumn{1}{c}{0.95} &
  \multicolumn{1}{c}{0.96} &
   \\ 
2 &
  \multicolumn{1}{c}{0.90} &
  \multicolumn{1}{c}{0.99} &
  \multicolumn{1}{c}{0.94} &
   &
  \multicolumn{1}{c}{0.89} &
  \multicolumn{1}{c}{0.98} &
  \multicolumn{1}{c}{0.93} &
   \\ 
3 &
  \multicolumn{1}{c}{0.99} &
  \multicolumn{1}{c}{0.90} &
  \multicolumn{1}{c}{0.94} &
   &
  \multicolumn{1}{c}{0.98} &
  \multicolumn{1}{c}{0.89} &
  \multicolumn{1}{c}{0.93} &
   \\ \hline
\end{tabular}%
}
\end{table}

\begin{table}[ht]
\centering
\caption{Classification report of the sequential Keras model for Dragon Board 1 for three (03) activities. The data set was split with a 70:30 ratio to training (70\%) and testing (30\%) of the model ( 0-Print Task, 1-Math Task, 2-Memory Task ).}
\label{tab:DB1_3 activity}
\resizebox{\textwidth}{!}{%
\def\arraystretch{1.2}
\begin{tabular}{ccccccccc}
\hline
 &
  \multicolumn{4}{c}{\textbf{Traning Data}} &
  \multicolumn{4}{c}{\textbf{Testing Data}} \\ \hline
\textbf{label} &
  \multicolumn{1}{c}{\textbf{precision}} &
  \multicolumn{1}{c}{\textbf{recall}} &
  \multicolumn{1}{c}{\textbf{F1-score}} &
  \textbf{accuracy} &
  \multicolumn{1}{c}{\textbf{precision}} &
  \multicolumn{1}{c}{\textbf{recall}} &
  \multicolumn{1}{c}{\textbf{F1-score}} &
  \textbf{accuracy} \\ \hline
0 &
  \multicolumn{1}{c}{0.98} &
  \multicolumn{1}{c}{0.94} &
  \multicolumn{1}{c}{0.96} &
  \multirow{3}{*}{0.97} &
  \multicolumn{1}{c}{0.97} &
  \multicolumn{1}{c}{0.93} &
  \multicolumn{1}{c}{0.95} &
  \multirow{3}{*}{0.96} \\ 
1 &
  \multicolumn{1}{c}{0.94} &
  \multicolumn{1}{c}{0.98} &
  \multicolumn{1}{c}{0.96} &
   &
  \multicolumn{1}{c}{0.93} &
  \multicolumn{1}{c}{0.96} &
  \multicolumn{1}{c}{0.95} &
   \\ 
2 &
  \multicolumn{1}{c}{0.99} &
  \multicolumn{1}{c}{1.00} &
  \multicolumn{1}{c}{1.00} &
   &
  \multicolumn{1}{c}{0.99} &
  \multicolumn{1}{c}{1.00} &
  \multicolumn{1}{c}{0.99} &
   \\ \hline
\end{tabular}%
}
\end{table}


\begin{table}[ht]
\centering
\caption{Classification report of the sequential Keras model for Dragon Board 1 for two(02) activities. The data set was split with a 70:30 ratio to training (70\%) and testing (30\%) the model ( 0-Math Task, 1-Memory Task ).}
\label{tab:DB1_2 activity}
\resizebox{\textwidth}{!}{%
\def\arraystretch{1.2}
\begin{tabular}{ccccccccc}
\hline
 &
  \multicolumn{4}{c}{\textbf{Traning Data}} &
  \multicolumn{4}{c}{\textbf{Testing Data}} \\ \hline
\textbf{label} &
  \multicolumn{1}{c}{\textbf{precision}} &
  \multicolumn{1}{c}{\textbf{recall}} &
  \multicolumn{1}{c}{\textbf{F1-score}} &
  \textbf{accuracy} &
  \multicolumn{1}{c}{\textbf{precision}} &
  \multicolumn{1}{c}{\textbf{recall}} &
  \multicolumn{1}{c}{\textbf{F1-score}} &
  \textbf{accuracy} \\ \hline
0 &
  \multicolumn{1}{c}{1.00} &
  \multicolumn{1}{c}{0.99} &
  \multicolumn{1}{c}{0.99} &
  \multirow{2}{*}{0.99} &
  \multicolumn{1}{c}{1.00} &
  \multicolumn{1}{c}{0.99} &
  \multicolumn{1}{c}{0.99} &
  \multirow{2}{*}{0.99} \\ 
1 &
  \multicolumn{1}{c}{0.99} &
  \multicolumn{1}{c}{1.00} &
  \multicolumn{1}{c}{0.99} &
   &
  \multicolumn{1}{c}{0.99} &
  \multicolumn{1}{c}{1.00} &
  \multicolumn{1}{c}{0.99} &
   \\ \hline
\end{tabular}%
}
\end{table}

\subsection{Challenges of using machine learning techniques in analyzing EM-SCA data in the crossed-IoT device portability context}

 In~\textbf{EXP1} Amazon Echo Show device, captured data belonging to four (04) activities. The final dataset contains 120,000 data samples for each amazon show device (cf. Table~\ref{tab:dataset}). Table~\ref{tab:echoshow} shows the classification report of the model build using the Echo Show device 1 dataset. This classifier achieved 91.0\% training accuracy, and testing accuracy of 89.0\% with acceptable precision, recall, and F1 score for the same device (Echo Dot 1) data used to train the model. There is a significant difference between the training and testing accuracy achieved by our classifier for the Echo Show 1 device (91\% and 89\%, respectively) compared to the accuracy achieved by the baseline study (99.66\%). In contrast, our Echo Show 1 model test with the Echo Show device 2 data shown in Table~\ref{tab:echoshow2} archived test accuracy of 25.0\% with extremely low precision, recall, and F1 scores for most of the classes.
  These results suggest that the model trained on one device may not generalize well to other devices, highlighting the need for further analysis, which was carried out in Experiment 2 (\textbf{EXP2}).

\begin{table}[!ht]
\centering
\caption{Classification report of the  Amazon Echo Show 1 model test with the  Amazon Echo Show 2 dataset ( 0-asking to play music, 1-asking to play radio, 2-asking for time, 3-asking for weather today ).}
\label{tab:echoshow2}
{%
\def\arraystretch{1.2}
\begin{tabular}{lcccc}
\hline
\multicolumn{1}{c}{} & \multicolumn{4}{c}{\textbf{Testing Data (crossed-devices)}}                                                                \\ \hline
\multicolumn{1}{c}{\textbf{label}} &
  \multicolumn{1}{c}{\textbf{precision}} &
  \multicolumn{1}{c}{\textbf{recall}} &
  \multicolumn{1}{c}{\textbf{F1-score}} &
  \textbf{accuracy} \\ \hline
0                      & \multicolumn{1}{c}{0.05} & \multicolumn{1}{c}{0.00} & \multicolumn{1}{c}{0.00} & \multirow{4}{*}{0.25} \\ 
1                      & \multicolumn{1}{c}{0.33} & \multicolumn{1}{c}{0.00} & \multicolumn{1}{c}{0.00} &                       \\ 
2                      & \multicolumn{1}{c}{0.25} & \multicolumn{1}{c}{1.00} & \multicolumn{1}{c}{0.40} &                       \\ 
3                      & \multicolumn{1}{c}{0.07} & \multicolumn{1}{c}{0.00} & \multicolumn{1}{c}{0.00} &                       \\ \hline
\end{tabular}%
}
\end{table}

\textbf{EXP2} aimed to further investigate the issues identified in~\textbf{EXP1} and to identify potential device variability. Dragon Boards were used as the Device Under Test (DUT), and separate Print, Math, Memory, and I/O activities were run on it while collecting CPU EM traces. Our hypothesis is that using multiple cores in SoC architecture can affect EM emissions and make it more difficult to identify the right source of emissions for a particular activity (cf. research question R1, Section \ref{sec:methodology}). Therefore, it affects the quality of EM data captured and as a consequence the overall accuracy of ML model dropped significantly (cf. Table~\ref{tab:echoshow2}). To prove this hypothesis, in \textbf{EXP2}, a single core of DUT is used for all activities in the experiment. The objective is to isolate the EM emissions from each activity. Another hypothesis is the number of activities, which is also a factor that affects the ML model's accuracy (cf. research question R2, Section \ref{sec:methodology}). To prove it, in \textbf{EXP2} we varied the number of activities in three (03) tasks as follows:

\begin{itemize}
    \item \textbf{Task 1 :} Involved building a model for Dragon Board 1 for four (04) activities: Print, Math, Memory, and I/O. The model achieved 96\% of training accuracy and testing accuracy of 95\%  with better precision, recall, and F1 score on the training dataset (cf. Table~\ref{tab:DB1_4 activity}). The model was tested with the Dragon Board 2 dataset for the same four activities. The model achieved an accuracy of 38\% on the test dataset shown in Table~\ref{tab:DB2_4 activity}. 
\\
    \item \textbf{Task 2 :} Involved building a model for Dragon Board 1 for three (03) activities: Print, Math, and Memory. The model achieved 97\% of training accuracy and testing accuracy of 96\%  with better precision, recall, and F1 score on the training dataset(cf. Table~\ref{tab:DB1_3 activity}). The model was tested with the Dragon Board 2 dataset for the same three activities. The model achieved an accuracy of 52\% on the test dataset shown in Table~\ref{tab:DB2_3 activity}. The final dataset used for this task contained 90,000 data points.
\\
    \item \textbf{Task 3 :} Involved building a model for Dragon Board 1 for two (02) activities: Math Task and Memory Task. The final dataset used for this task contained 60,000 data points. The model achieved training and testing accuracy of 99\% with acceptable precision, recall, and F1 score on the training dataset(cf. Table~\ref{tab:DB1_2 activity}). The Dragon Board 1 model was tested with the Dragon Board 2 dataset for the same two activities, Math and Memory tasks, and the accuracy dropped to 77\% on the test dataset shown in Table~\ref{tab:DB2_2 activity}.
\end{itemize}


\begin{table}[!ht]
\centering
\caption{Classification report of the Dragon Board 1 model test with the Dragon Board 2 dataset for four(04) activities. ( 0-Print Task, 1-Math Task, 2-Memory Task, 3-I/O Task )}
\label{tab:DB2_4 activity}
{%
\def\arraystretch{1.2}
\begin{tabular}{ccccc}
\hline
  & \multicolumn{4}{c}{\textbf{Testing Data (crossed-devices)}}                                           \\ \hline
\textbf{label} & \multicolumn{1}{c}{\textbf{precision}} & \multicolumn{1}{c}{\textbf{recall}} & \multicolumn{1}{c}{\textbf{F1-score}} & \textbf{accuracy}     \\ \hline
0              & \multicolumn{1}{c}{0.72}               & \multicolumn{1}{c}{0.09}            & \multicolumn{1}{c}{0.16}              & \multirow{4}{*}{0.38} \\ 
1 & \multicolumn{1}{c}{0.38} & \multicolumn{1}{c}{0.85} & \multicolumn{1}{c}{0.52} &  \\ 
2 & \multicolumn{1}{c}{0.35} & \multicolumn{1}{c}{0.57} & \multicolumn{1}{c}{0.44} &  \\ 
3 & \multicolumn{1}{c}{0.85} & \multicolumn{1}{c}{0.01} & \multicolumn{1}{c}{0.02} &  \\ \hline
\end{tabular}%
}
\end{table}


\begin{table}[ht]
\centering
\caption{Classification report of the Dragon Board 1 model test with the Dragon Board 2 dataset for three (03) activities.(0-Print Task, 1-Math Task, 2-Memory Task)}
\label{tab:DB2_3 activity}
{%
\def\arraystretch{1.2}
\begin{tabular}{ccccc}
\hline
  & \multicolumn{4}{c}{\textbf{Testing Data (crossed-devices)}}                                            \\ \hline
\textbf{label} & \multicolumn{1}{c}{\textbf{precision}} & \multicolumn{1}{c}{\textbf{recall}} & \multicolumn{1}{c}{\textbf{F1-score}} & \textbf{accuracy} \\ \hline
0 & \multicolumn{1}{c}{0.70} & \multicolumn{1}{c}{0.05} & \multicolumn{1}{c}{0.10} & \multirow{3}{*}{0.52} \\ 
1 & \multicolumn{1}{c}{0.42} & \multicolumn{1}{c}{0.93} & \multicolumn{1}{c}{0.58} &                       \\ 
2 & \multicolumn{1}{c}{0.82} & \multicolumn{1}{c}{0.56} & \multicolumn{1}{c}{0.67} &                       \\ \hline
\end{tabular}%
}
\end{table}


\begin{table}[ht]
\centering
\caption{Classification report of the Dragon Board 1 model test with the Dragon Board 2 dataset for two(02) activities. ( 0-Math Task, 1-Memory Task )}
\label{tab:DB2_2 activity}
{%
\def\arraystretch{1.2}
\begin{tabular}{ccccc}
\hline
  & \multicolumn{4}{c}{\textbf{Testing Data (crossed-devices)}}                                            \\ \hline
\textbf{label} & \multicolumn{1}{c}{\textbf{precision}} & \multicolumn{1}{c}{\textbf{recall}} & \multicolumn{1}{c}{\textbf{F1-score}} & \textbf{accuracy} \\ \hline
0 & \multicolumn{1}{c}{0.70} & \multicolumn{1}{c}{0.93} & \multicolumn{1}{c}{0.80} & \multirow{2}{*}{0.77} \\ 
1 & \multicolumn{1}{c}{0.89} & \multicolumn{1}{c}{0.61} & \multicolumn{1}{c}{0.73} &                       \\ \hline
\end{tabular}%
}
\end{table}

\subsection{Analyzing the feasibility of using transfer learning}

Furthermore, to address the last research question R3 (cf. Section\ref{sec:methodology}) we employed transfer learning techniques to enhance the performance of our models developed on the Echo Show 1 and Dragon Board 1 devices. Transfer learning is a widely used approach in machine learning that enables us to leverage the knowledge gained from one task to improve the performance of another related task~\cite{zhuang2020comprehensive}. Specifically, we utilized a pre-trained model as the starting point and fine-tuned it on a new dataset or task, which allowed the model to benefit from the features learned by the pre-trained model and achieve superior performance on new datasets. To achieve this, we utilized a pre-trained model and froze the weights of the pre-trained layers up to the first dense layer. Subsequently, we added new layers on top of the pre-trained model, such as a transfer dense layer and an output layer, to build a new Keras model. We then trained this model on the respective data sets for each device (Echo Show 2 and Dragon Board 2) in our experiments. We repeated this process for all three models developed in our study. 


\begin{table}[ht]
\centering
\caption{Classification report of the \textbf{transfer learning} Keras model for the Amazon Echo Show 2 dataset, after fine-tuning the pre-trained Amazon Echo Show 1 model. The data set was split with a 70:30 ratio to training (70\%) and testing (30\%) the model ( 0-asking to play music, 1-asking to play radio, 2-asking for time, 3-asking for weather today ).}
\label{tab:trasfer_echoshow}
\resizebox{\textwidth}{!}{%
\def\arraystretch{1.2}
\begin{tabular}{ccccccccc}
\hline
 &
  \multicolumn{4}{c}{\textbf{Traning Data}} &
  \multicolumn{4}{c}{\textbf{Testing Data}} \\ \hline
\textbf{label} &
  \multicolumn{1}{c}{\textbf{precision}} &
  \multicolumn{1}{c}{\textbf{recall}} &
  \multicolumn{1}{c}{\textbf{F1-score}} &
  \textbf{accuracy} &
  \multicolumn{1}{c}{\textbf{precision}} &
  \multicolumn{1}{c}{\textbf{recall}} &
  \multicolumn{1}{c}{\textbf{F1-score}} &
  \textbf{accuracy} \\ \hline
0 &
  \multicolumn{1}{c}{0.56} &
  \multicolumn{1}{c}{0.59} &
  \multicolumn{1}{c}{0.58} &
  \multirow{4}{*}{0.53} &
  \multicolumn{1}{c}{0.56} &
  \multicolumn{1}{c}{0.59} &
  \multicolumn{1}{c}{0.57} &
  \multirow{4}{*}{0.53} \\ 
1 &
  \multicolumn{1}{c}{0.46} &
  \multicolumn{1}{c}{0.38} &
  \multicolumn{1}{c}{0.42} &
   &
  \multicolumn{1}{c}{0.46} &
  \multicolumn{1}{c}{0.38} &
  \multicolumn{1}{c}{0.42} &
   \\ 
2 &
  \multicolumn{1}{c}{0.65} &
  \multicolumn{1}{c}{0.70} &
  \multicolumn{1}{c}{0.67} &
   &
  \multicolumn{1}{c}{0.65} &
  \multicolumn{1}{c}{0.69} &
  \multicolumn{1}{c}{0.67} &
   \\ 
3 &
  \multicolumn{1}{c}{0.44} &
  \multicolumn{1}{c}{0.47} &
  \multicolumn{1}{c}{0.46} &
   &
  \multicolumn{1}{c}{0.44} &
  \multicolumn{1}{c}{0.46} &
  \multicolumn{1}{c}{0.45} &
   \\ \hline
\end{tabular}%
}
\end{table}

For Echo Show 2 data, we significantly improved our models' accuracy by using a transfer learning approach, with training and testing accuracy reaching 53\% for both shown in Table~\ref{tab:trasfer_echoshow}. Previously, Echo Show 2 data was archived with only 25\% accuracy when tested on the Echo Show 1 model. This result demonstrates the effectiveness of transfer learning in improving model performance on new devices. Similarly, for Dragon Board 2, we achieved remarkable improvements in the accuracy of our models by using transfer learning techniques. New model archived 57\% training and testing accuracy (cf. Table~\ref{tab:trasfer_DB2_4activity}) for the four (04) activities run on the Dragon Board 2. The model improved from 38\% tested using the Dragon Board 1 model (cf. Table \ref{tab:DB2_4 activity}). Although, the training and testing accuracy for the three activities reached 68\% and 67\%, respectively (cf. Table \ref{tab:trasfer_DB2_3activity}). Previously, Dragon Board 2 data was archived with only 52\% accuracy when tested on the Dragon Board 1 model. We also achieved high accuracy of 88\% and 87\% for two activities on Dragon Board 2, which improved from 77\% tested using the Dragon Board 1 model (cf. Table \ref{tab:trasfer_DB2_2activity}). Our approach of fine-tuning pre-trained models and adding new layers on top allowed us to extract valuable features from pre-trained models and adapt them to new tasks, resulting in superior performance. 


\begin{table}[ht]
\centering
\caption{Classification report of the \textbf{transfer learning} Keras model for the Dragon Board 2 dataset, after fine-tuning the pre-trained Dragon Board 1 model. The data set was split with a 70:30 ratio to training (70\%) and testing (30\%) of the model ( 0-Print Task, 1-Math Task, 2-Memory Task, 3-I/O Task )}
\label{tab:trasfer_DB2_4activity}
\resizebox{\textwidth}{!}{%
\def\arraystretch{1.2}
\begin{tabular}{ccccccccc}
\hline
 &
  \multicolumn{4}{c}{\textbf{Traning Data}} &
  \multicolumn{4}{c}{\textbf{Testing Data}} \\ \hline
\textbf{label} &
  \multicolumn{1}{c}{\textbf{precision}} &
  \multicolumn{1}{c}{\textbf{recall}} &
  \multicolumn{1}{c}{\textbf{F1-score}} &
  \textbf{accuracy} &
  \multicolumn{1}{c}{\textbf{precision}} &
  \multicolumn{1}{c}{\textbf{recall}} &
  \multicolumn{1}{c}{\textbf{F1-score}} &
  \textbf{accuracy} \\ \hline
0 &
  \multicolumn{1}{c}{0.53} &
  \multicolumn{1}{c}{0.54} &
  \multicolumn{1}{c}{0.54} &
  \multirow{4}{*}{0.57} &
  \multicolumn{1}{c}{0.53} &
  \multicolumn{1}{c}{0.55} &
  \multicolumn{1}{c}{0.54} &
  \multirow{4}{*}{0.57} \\ 
1 &
  \multicolumn{1}{c}{0.62} &
  \multicolumn{1}{c}{0.60} &
  \multicolumn{1}{c}{0.61} &
   &
  \multicolumn{1}{c}{0.62} &
  \multicolumn{1}{c}{0.60} &
  \multicolumn{1}{c}{0.61} &
   \\ 
2 &
  \multicolumn{1}{c}{0.53} &
  \multicolumn{1}{c}{0.44} &
  \multicolumn{1}{c}{0.48} &
   &
  \multicolumn{1}{c}{0.53} &
  \multicolumn{1}{c}{0.43} &
  \multicolumn{1}{c}{0.47} &
   \\ 
3 &
  \multicolumn{1}{c}{0.59} &
  \multicolumn{1}{c}{0.69} &
  \multicolumn{1}{c}{0.63} &
   &
  \multicolumn{1}{c}{0.58} &
  \multicolumn{1}{c}{0.69} &
  \multicolumn{1}{c}{0.63} &
   \\ \hline
\end{tabular}%
}
\end{table}

\begin{table}[!ht]
\centering
\caption{Classification report of the \textbf{transfer learning} Keras model for the Dragon Board 2 dataset, after fine-tuning the pre-trained Dragon Board 1 model. The data set was split with a 70:30 ratio to training (70\%) and testing (30\%) of the model ( 0-Print Task, 1-Math Task, 2-Memory Task ).}
\label{tab:trasfer_DB2_3activity}
\resizebox{\textwidth}{!}{%
\def\arraystretch{1.2}
\begin{tabular}{ccccccccc}
\hline
 &
  \multicolumn{4}{c}{\textbf{Traning Data}} &
  \multicolumn{4}{c}{\textbf{Testing Data}} \\ \hline
\textbf{label} &
  \multicolumn{1}{c}{\textbf{precision}} &
  \multicolumn{1}{c}{\textbf{recall}} &
  \multicolumn{1}{c}{\textbf{F1-score}} &
  \textbf{accuracy} &
  \multicolumn{1}{c}{\textbf{precision}} &
  \multicolumn{1}{c}{\textbf{recall}} &
  \multicolumn{1}{c}{\textbf{F1-score}} &
  \textbf{accuracy} \\ \hline
0 &
  \multicolumn{1}{c}{0.60} &
  \multicolumn{1}{c}{0.50} &
  \multicolumn{1}{c}{0.55} &
  \multirow{3}{*}{0.68} &
  \multicolumn{1}{c}{0.60} &
  \multicolumn{1}{c}{0.52} &
  \multicolumn{1}{c}{0.56} &
  \multirow{3}{*}{0.67} \\
1 &
  \multicolumn{1}{c}{0.63} &
  \multicolumn{1}{c}{0.71} &
  \multicolumn{1}{c}{0.67} &
   &
  \multicolumn{1}{c}{0.64} &
  \multicolumn{1}{c}{0.70} &
  \multicolumn{1}{c}{0.67} &
   \\ 
2 &
  \multicolumn{1}{c}{0.77} &
  \multicolumn{1}{c}{0.81} &
  \multicolumn{1}{c}{0.79} &
   &
  \multicolumn{1}{c}{0.77} &
  \multicolumn{1}{c}{0.81} &
  \multicolumn{1}{c}{0.79} &
   \\ \hline
\end{tabular}%
}
\end{table}


\begin{table}[ht]
\centering
\caption{Classification report of the \textbf{transfer learning} Keras model for the Dragon Board 2 dataset, after fine-tuning the pre-trained Dragon Board 1 model. The data set was split with a 70:30 ratio to training (70\%) and testing (30\%) the model ( 0-MathTask, 1-Memory Task ).}
\label{tab:trasfer_DB2_2activity}
\resizebox{\textwidth}{!}{%
\def\arraystretch{1.2}
\begin{tabular}{ccccccccc}
\hline
 &
  \multicolumn{4}{c}{\textbf{Traning Data}} &
  \multicolumn{4}{c}{\textbf{Testing Data}} \\ \hline
\textbf{label} &
  \multicolumn{1}{c}{\textbf{precision}} &
  \multicolumn{1}{c}{\textbf{recall}} &
  \multicolumn{1}{c}{\textbf{F1-score}} &
  \textbf{accuracy} &
  \multicolumn{1}{c}{\textbf{precision}} &
  \multicolumn{1}{c}{\textbf{recall}} &
  \multicolumn{1}{c}{\textbf{F1-score}} &
  \textbf{accuracy} \\ \hline
0 &
  \multicolumn{1}{c}{0.86} &
  \multicolumn{1}{c}{0.90} &
  \multicolumn{1}{c}{0.88} &
  \multirow{2}{*}{0.88} &
  \multicolumn{1}{c}{0.86} &
  \multicolumn{1}{c}{0.89} &
  \multicolumn{1}{c}{0.87} &
  \multirow{2}{*}{0.87} \\ 
1 &
  \multicolumn{1}{c}{0.89} &
  \multicolumn{1}{c}{0.86} &
  \multicolumn{1}{c}{0.87} &
   &
  \multicolumn{1}{c}{0.89} &
  \multicolumn{1}{c}{0.85} &
  \multicolumn{1}{c}{0.87} &
   \\ \hline
\end{tabular}%
}
\end{table}

The results showed that the performance of the models varied significantly depending on the device and the data used to train the models. We notices that the device variability is directly affected on the model performance in Electromagnetic Side Channel Analysis datasets we used in our experiments. Further, we identified environmental factors can significantly affect the electromagnetic emissions of devices and, therefore, the quality of the data collected. Although, transfer learning is effective technique for analyse Electromagnetic Side Channel Analysis datasets and overcome device variability problem.

\section{Discussion}
\label{sec:discussion}

\begin{figure}[!ht]
    \centering
    \includegraphics[width= 12cm]{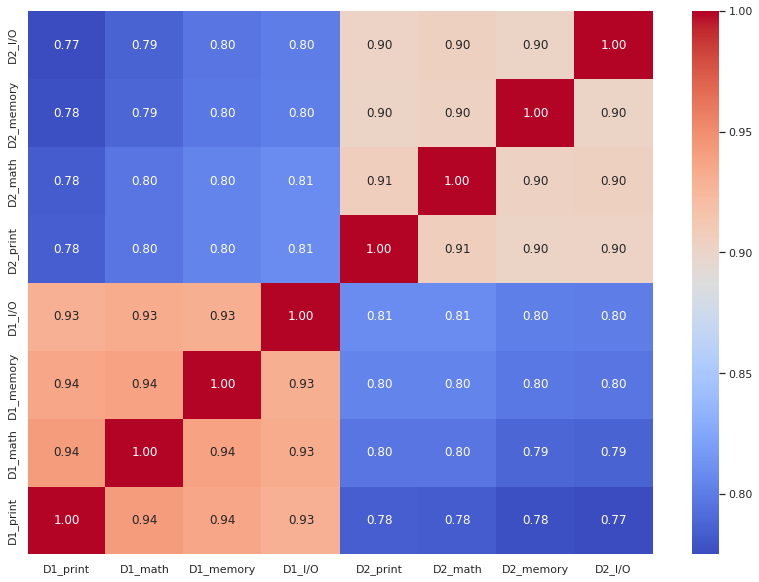}
    \caption{A heat map was generated to perform correlation analysis for activities on Dragon Board 1(D1) and Dragon Board 2(D2). Only one core (core 1) was utilized for all the activities on both Dragon Boards.}
    \label{fig:db_corr}
\end{figure}

The results of our study demonstrate that ML methods can effectively analyze EM-SCA data to identify the software activities of smart devices. In addition, our study also highlights the limitations of the ML-based approaches, including the need for careful data preprocessing and the potential impact of device variability and environmental factors on the model performance.

In this research, we identified device variability significantly impacts the model's performance in EM-SCA datasets. As a result, we encountered low precision, recall, and F1 scores obtained when testing the models with different devices' data. Furthermore, the low precision, recall, and F1 scores obtained in our study are significant in understanding the overall accuracy of the models. It is essential to highlight that low precision means that the model identifies many false positives. In other words, the model incorrectly classifies data from one activity as belonging to another. Similarly, low recall means that the model needs many true positives. In other words, the model is not recognizing data from one activity as belonging to that activity. Lastly, a low F1 score means that the model has an overall low accuracy in correctly identifying data belonging to a specific activity. In contrast, the results obtained using our experiments significantly differ from the baseline study we used to validate our results. Although, we assume the experimental setup we used caused the outcomes in our study. In the baseline study~\cite{sayakkara2021electromagnetic}, authors captured electromagnetic traces directly from the Echo Dot/ Echo Show/ Smart devices, while we used a Dragonboard and Echo Show as a DUT in~\textbf{EXP1} to capture the network traffic through the Dragonboard as the EM traces. This difference in the data source could have led to variations in the quality and quantity of the captured data, which could impact the classification accuracy, precision, recall, and F1 scores. 


\begin{figure}[ht]
    \centering
    \includegraphics[width= 12cm]{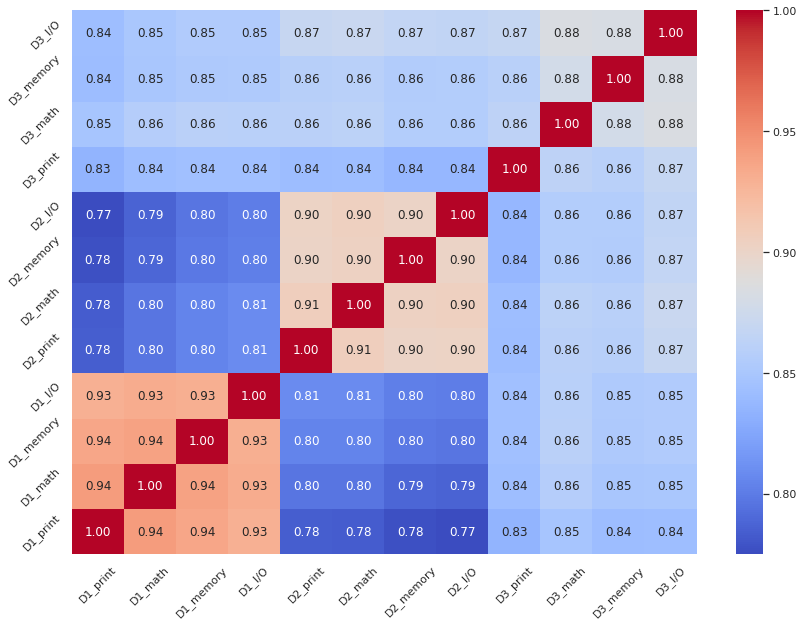}
    \caption{A heat map was generated to perform correlation analysis for activities on Dragon Board 1(D1), Dragon Board 2(D2) and Dragon Board 3(D3). Only one core (core 1) was utilized for all the activities on all Dragon Boards.}
    \label{fig:db_corr_3d}
\end{figure}

Besides, most smart devices use multi-core CPUs to divide the workload and complete tasks more quickly. Using multiple cores can affect electromagnetic (EM) emissions, as the increased processing power can generate more electromagnetic interference (EMI). This interference can make it more difficult to identify the source of emissions for a particular activity, as the signals from multiple cores may overlap and create a complex and unpredictable electromagnetic environment. In \textbf{EXP2} we used only one (1) core out of four (4) cores in the Dragon Board CPU and run different activities to investigate these problems. The results of~\textbf{EXP2} showed a significant drop in accuracy when models trained on one device were tested on another, even if they were of the same model. In contrast, we observed a high correlation between the different activities, as the EM emissions from one activity can be mistaken for another (cf. Figure~\ref{fig:db_corr}). As the number of activities increases, the correlation between them also increases, making distinguishing between activities based on EM emissions more challenging. In contrast, if the number of devices increase may lead to a high correlation problem between similar activities (cf. Figure~\ref{fig:db_corr_3d}). This can lead to more false positives, where the model identifies an activity as present when it is not. The higher number of false positives can further reduce the model's accuracy. Additionally, the experiment demonstrated that device variability significantly impacts the accuracy of the models, and testing the models on datasets from multiple devices is essential to ensure their robustness.

Despite these differences, we still found the baseline study~\cite{sayakkara2021electromagnetic} to be a valuable point of comparison for our results. This outcome suggests that the models built with one device's data may not generalize well to other devices, and their accuracy may significantly decrease. The summary of results shows that EM-SCA datasets can vary significantly between devices, and therefore, it is essential to consider device variability when analyzing EM data. This finding is significant in practical applications where the goal is to identify and classify the activity of a device in a diverse range of real-world scenarios. 

Our study also identified that environmental factors significantly affect the electromagnetic emissions of devices and the quality of the data collected. The low precision, recall, and F1 scores in these cases highlight the importance of training the models with data representative of the device and its operating environment. Any changes in the device or its environment can significantly affect the electromagnetic emissions of the device, leading to data that may not be representative of the device's normal operation. So, environmental factors also affect models with low accuracy, as seen in our study. Further investigation is needed to optimize our data collection and preprocessing methods to improve the accuracy, precision, recall, and F1 scores of our classifier.

One limitation of our study is that we focused on a limited set of smart devices and software activities. Future research could explore the performance of machine learning models on a broader range of devices and activities and the impact of other factors, such as device age and usage patterns, on model performance. Another limitation of smart devices is the lack of control over the hardware level of each device. For example, the Echo Show has a 4-core CPU, and difficult to configure it to use a specific core, such as the Dragon board. This can have consequences in terms of electromagnetic (EM) emissions, as increased processing power can generate more electromagnetic interference (EMI). This makes more difficult to identify the source of emissions for a particular activity. With multiple cores in a CPU, signals can overlap and create a complex and unpredictable electromagnetic environment.
The lack of control over the hardware level of smart devices can limit the ability to mitigate any potential electromagnetic interference and cause to the device variability problem.

Transfer learning involves using pre-trained models on similar tasks to improve the performance of a new model. In the context of our study, transfer learning shows promising outcomes on EM trace data from similar smart devices to improve the accuracy and generalizability of the models developed in this work. Also, Transfer learning could be a promising approach for the future use of this work. Another potential direction for future research is to explore other data preprocessing techniques, such as wavelet transforms\cite{zhang2019wavelet}, to improve feature extraction from EM trace data. In addition, further investigation is needed to identify the environmental factors that can affect the quality of the EM trace data and how to mitigate these effects in data collection and analysis.

Based on the analysis presented in our research paper, several new research directions have been identified,
\begin{itemize}
    \item \textbf{Multi-core problem for EM-SCA data:} This is an area of research that could explore ways to optimize the processing of EMSCA data using multi-core architectures. Additionally, an important direction to investigate the impact of using multiple cores on EM emissions and how this affects the identification of the source of emissions. The number of cores in smart device CPU is potentially significant factor that may affect the accuracy of current EM analysis techniques.
    \\
    \item \textbf{High correlation between similar activities in different devices from the same model:} This could be another interesting research direction that examines the extent of correlation between activities across devices of the same model. It leads to insights into the consistency of sensor measurements across different devices and potentially inform efforts to improve data quality. It highlights the need for more extensive data collection and analysis to accurately identify the source of EM emissions.
    \\
    \item \textbf{Transfer learning methods:} The study highlighted the potential for transfer learning to improve activity recognition performance, particularly in situations where training data not performing well on the ML models. Future research could focus on exploring the effectiveness of transfer learning in different scenarios, as well as developing new transfer learning techniques that are tailored to specific applications.
\end{itemize}

Overall, our study highlights the potential of machine learning for analyzing EM trace data in the context of side-channel analysis while also underscoring the need for careful consideration of data preprocessing, device variability, and environmental factors in model development and evaluation.

\section{Conclusion}
\label{sec:conclusion}
In conclusion, the growth of IoT devices has led to the emergence of IoT forensics, which plays a critical role in investigating and preventing malicious activities on IoT devices. EM-SCA has become an essential tool for IoT forensics due to its ability to reveal confidential information about the internal workings of IoT devices. However, the accuracy and reliability of EM-SCA results can be limited by several factors, including device variability, environmental factors, and data collection and processing methods. Our study provides insights into the challenges of developing machine learning models for the side-channel analysis of smart devices. Device variability and environmental factors may significantly affect the performance of those models.

Further research is required to improve the accuracy and reliability of EM-SCA results for IoT forensics and to overcome the limitations and challenges associated with EM-SCA datasets. This study provides a foundation for future work in this area. It presents an opportunity for the research community to address some critical issues associated with using EM-SCA in IoT forensics.

\bibliographystyle{unsrt}  
\bibliography{templateArxiv}

\begin{thebibliography}{10}

\bibitem{gokhale2018introduction}
Pradyumna Gokhale, Omkar Bhat, and Sagar Bhat.
\newblock Introduction to iot.
\newblock {\em International Advanced Research Journal in Science, Engineering and Technology}, 5(1):41--44, 2018.

\bibitem{atlam2020internet}
Hany~F Atlam, Ezz El-Din Hemdan, Ahmed Alenezi, Madini~O Alassafi, and Gary~B Wills.
\newblock Internet of things forensics: A review.
\newblock {\em Internet of Things}, 11:100220, 2020.

\bibitem{conti2018internet}
Mauro Conti, Ali Dehghantanha, Katrin Franke, and Steve Watson.
\newblock Internet of things security and forensics: Challenges and opportunities, 2018.

\bibitem{ghosh2022electromagnetic}
Archisman Ghosh, Mayukh Nath, Debayan Das, Santosh Ghosh, and Shreyas Sen.
\newblock Electromagnetic analysis of integrated on-chip sensing loop for side-channel and fault-injection attack detection.
\newblock {\em IEEE Microwave and Wireless Components Letters}, 32(6):784--787, 2022.

\bibitem{sayakkara2020facilitating}
Asanka Sayakkara, Nhien-An Le-Khac, and Mark Scanlon.
\newblock Facilitating electromagnetic side-channel analysis for iot investigation: Evaluating the emvidence framework.
\newblock {\em Forensic Science International: Digital Investigation}, 33:301003, 2020.

\bibitem{sayakkara2019survey}
Asanka Sayakkara, Nhien-An Le-Khac, and Mark Scanlon.
\newblock A survey of electromagnetic side-channel attacks and discussion on their case-progressing potential for digital forensics.
\newblock {\em Digital Investigation}, 29:43--54, 2019.

\bibitem{KAMILARIS201870}
Andreas Kamilaris and Francesc~X. Prenafeta-Boldú.
\newblock Deep learning in agriculture: A survey.
\newblock {\em Computers and Electronics in Agriculture}, 147:70--90, 2018.

\bibitem{SriRevathi2023}
Sri~Revathi B.
\newblock A survey on advanced machine learning and deep learning techniques assisting in renewable energy generation.
\newblock {\em Environ Sci Pollut Res}, 30:93407–93421, 2023.

\bibitem{ZENNARO2021103752}
Federica Zennaro, Elisa Furlan, Christian Simeoni, Silvia Torresan, Sinem Aslan, Andrea Critto, and Antonio Marcomini.
\newblock Exploring machine learning potential for climate change risk assessment.
\newblock {\em Earth-Science Reviews}, 220:103752, 2021.

\bibitem{Bhardwaj2017}
R.~Bhardwaj, A.~R. Nambiar, and D.~Dutta.
\newblock A study of machine learning in healthcare.
\newblock In {\em 2017 IEEE 41st Annual Computer Software and Applications Conference}, pages 236--241, Turin, Italy, July 2017.

\bibitem{Shaukat2020}
Kamran Shaukat, Suhuai Luo, Vijay Varadharajan, Ibrahim~A. Hameed, and Min Xu.
\newblock A survey on machine learning techniques for cyber security in the last decade.
\newblock {\em IEEE Access}, 8:222310--222354, 2020.

\bibitem{Shahzad2020}
Faisal Shahzad, Abdul~Rehman Javed, Zunera Jalil, and Farkhund Iqbal.
\newblock {\em Cyber Forensics with Machine Learning}, pages 1--6.
\newblock Springer US, New York, NY, 2020.

\bibitem{Sayakkara2021smartphone}
Asanka~P. Sayakkara and Nhien-An Le-Khac.
\newblock Forensic insights from smartphones through electromagnetic side-channel analysis.
\newblock {\em IEEE Access}, 9:13237--13247, 2021.

\bibitem{sayakkara2021electromagnetic}
Asanka~P Sayakkara and Nhien-An Le-Khac.
\newblock Electromagnetic side-channel analysis for iot forensics: Challenges, framework, and datasets.
\newblock {\em IEEE Access}, 9:113585--113598, 2021.

\bibitem{hong2017extracting}
Seungwon Hong, Dae-Hyun Kim, and Hyunchul Lee.
\newblock Extracting encryption key from an iot device using electromagnetic side-channel analysis.
\newblock {\em Journal of Ambient Intelligence and Humanized Computing}, 8(1):129--142, 2017.

\bibitem{kim2018detecting}
Dae-Hyun Kim, Seungwon Hong, and Hyunchul Lee.
\newblock Detecting malware in iot devices using electromagnetic side-channel analysis.
\newblock {\em Journal of Ambient Intelligence and Humanized Computing}, 9(1):69--80, 2018.

\bibitem{zhao2019limitations}
Lu~Zhao, Xiaoyu Wang, Wei Li, and Mingzhe Gao.
\newblock Limitations in electromagnetic side-channel analysis datasets for iot forensics.
\newblock {\em Journal of Ambient Intelligence and Humanized Computing}, 10(1):55--66, 2019.

\bibitem{chen2020machine}
Jun Chen, Shaogang Lu, Wei Zhang, and Xiaoyu Wang.
\newblock Machine learning techniques in electromagnetic side-channel analysis for iot forensics.
\newblock {\em Journal of Ambient Intelligence and Humanized Computing}, 11(1):43--54, 2020.

\bibitem{levina2016side}
Alia Levina, Daria Sleptsova, and Oleg Zaitsev.
\newblock Side-channel attacks and machine learning approach.
\newblock In {\em 2016 18th Conference of Open Innovations Association and Seminar on Information Security and Protection of Information Technology (FRUCT-ISPIT)}, pages 181--186. IEEE, 2016.

\bibitem{zhuang2020comprehensive}
Fuzhen Zhuang, Zhiyuan Qi, Keyu Duan, Dongbo Xi, Yongchun Zhu, Hengshu Zhu, Hui Xiong, and Qing He.
\newblock A comprehensive survey on transfer learning.
\newblock {\em Proceedings of the IEEE}, 109(1):43--76, 2020.

\bibitem{zhang2019wavelet}
Dengsheng Zhang and Dengsheng Zhang.
\newblock Wavelet transform.
\newblock {\em Fundamentals of Image Data Mining: Analysis, Features, Classification and Retrieval}, pages 35--44, 2019.

\end{thebibliography}

\end{document}